\def\year{2021}\relax
\documentclass[letterpaper]{article} 

\usepackage{natbib}  
\usepackage[hyphens]{url}  
\usepackage{graphicx} 
\usepackage{microtype}
\usepackage[dvipsnames]{xcolor}
\usepackage{subfigure}
\usepackage{booktabs} 
\usepackage{amssymb}
\usepackage{todonotes}
\usepackage{mathtools}
\usepackage{multirow}
\usepackage{comment}
\usepackage{hyperref}
\usepackage{xspace}
\usepackage[switch]{lineno}  %

\DeclarePairedDelimiterX{\infdivx}[2]{(}{)}{%
#1\;\delimsize|\delimsize|\;#2%
}
\newcommand{\kl}[2]{\ensuremath{\text{KL} \infdivx{#1}{#2}}\xspace}

\DeclareMathOperator{\E}{\mathbb{E}}
\graphicspath{ {imgs/} }

\usepackage{aaai21}  
\usepackage{times}  
\usepackage{helvet} 
\usepackage{courier}  
\usepackage{caption} 
\title{A Study on the Autoregressive and non-Autoregressive Multi-label Learning }
\author{
    Elham J. Barezi\textsuperscript{\rm 1}
    Iacer Calixto\textsuperscript{\rm 2,3}
    Kyunghyun Cho\textsuperscript{\rm 2}
    Pascale Fung\textsuperscript{\rm 1}\\
}
\affiliations{

    \textsuperscript{\rm 1}Department of Computer Science and Engineering, The Hong Kong University of Science and Technology, Clear Water Bay, Hong Kong\\
    \textsuperscript{\rm 2}Courant Institute of Mathematical Sciences, New York University, US\\
    \textsuperscript{\rm 3}Institute for Logic, Language and Computation, The University of Amsterdam, NL\\
    \texttt{\href{mailto:ejs@ust.hk}{ejs@ust.hk}}, \texttt{\{iacer.calixto,kyunghyun.cho\}@nyu.edu}, \texttt{pascale@ust.hk}\\
}

\begin{document}
\maketitle

\begin{abstract}
\textit{Extreme classification} tasks are multi-label tasks with an extremely large number of labels (tags).
These tasks are hard because the label space is usually \textit{(i)} very large, e.g. thousands or millions of labels, \textit{(ii)} very sparse, i.e. very few labels apply to each input document, and \textit{(iii)} highly correlated, meaning that the existence of one label changes the likelihood of predicting all other labels.
In this work, we propose a self-attention based variational encoder-model to extract the label-label and label-feature dependencies jointly and to predict labels for a given input.
In more detail, we propose a non-autoregressive latent variable model and compare it to a strong autoregressive baseline that predicts a label based on all previously generated labels.
Our model can therefore be used to predict all labels \textit{in parallel while still including both label-label and label-feature dependencies through latent variables}, and compares favourably to the autoregressive baseline.
We apply our models to four standard extreme classification natural language data sets, and one news videos dataset for automated label detection from a lexicon of semantic concepts.
Experimental results show that although the autoregressive models, where use a given order of the labels for chain-order label prediction, work great for the small scale labels or the prediction of the  highly ranked label, but our non-autoregressive model surpasses them by around 2\% to 6\% when we need to predict more labels, or the dataset has a larger number of the labels.  
\end{abstract}
\section{Introduction}

Multi-label learning has recently attracted attention in the research community due to an increase in applications such as semantic labeling of images and videos \cite{qi2007correlative,wang2008automatic}, bioinformatics and genetic function \cite{zhang2006multilabel} and music categorization \cite{trohidis2008multi,sanden2011enhancing}, to name a few. Moreover, multi-label learning can address machine learning problems in web data mining, including recommendation systems, multimedia sharing websites, and ranking~\cite{zhang2010multi, zhang2013review,tsoumakas2009mining,babbar2019data}.

A particularly interesting subset of multi-label learning involve \textit{extreme classification} tasks, which consist of multi-label learning tasks with a very large number of labels or tags.
These tasks often exhibit strong label interdependence and very sparse label distribution, so that na\"{i}vely assuming independence in label space leads to rather weak models; on the other hand, because of the exponential number of possible label combinations directly modelling all label interdependencies is computationally intractable~\cite{zhang2010multi,tsoumakas2010mining,xu2019survey,babbar2019data}.
The range of tasks that fall under the ``extreme classification'' umbrella is very broad, and include applications such as social tagging of large web databases spanning text, images, videos, genomic data, etc.
Tags by definition are open vocabulary, and the number of tags tend to grow continually in order to adjust to the needs of new information and also due to the fact different users can assign different tags to the same resource (e.g. Wikipedia).

Classifier chains are a state-of-the-art method for tackling such problems, which essentially converts this problem into a sequential prediction problem, where the labels are first ordered in an arbitrary fashion, and the task is to predict a sequence of binary values for these labels. \cite{nam2017maximizing} replaced classifier chains with recurrent neural networks, a sequence-to-sequence (seq2seq) prediction algorithm which has been successfully applied to sequential prediction tasks in many domains. 
These tasks have been successfully tackled using sequence prediction framework \cite{wang2016cnn,gehring2017convolutional,nam2017maximizing}, and autoregressive (AR) models usually perform well in these problems \cite{nam2017maximizing}. They capture interdependence in the output space, can be efficiently trained using SGD, and generalize well enough for practical purposes. However, both classifier chains and recurrent neural networks depend on a fixed ordering of the labels, which is typically not part of a multi-label problem specification.
Moreover, inference with these models cannot be easily parallelized due to the autoregressive factorization of the output probability, and AR models tend to struggle with increased output sequence lengths.

In this work, we follow a recent trend in seq2seq learning \cite{gu2017non,lee2018deterministic,kaiser2018fast} and propose a \textit{non-autoregressive} latent variable model for extreme classification.
We demonstrate that our model compares favourably to their autoregressive counterparts while being fully paralellizable,
and we validate our method on datasets involving social tagging of large information collections spanning across text and video, such as Wikipedia.

Our findings indicate that the proposed transformer model outperforms the baseline seq2seq models using autoregressive structures for predicting more labels or working on the larger set of the labels, even if we put a deeper architecture on top of that model and give it strictly more parameters. We argue that this is due to the transformer’s ability to employ self-attention over both input and label spaces simultaneously.

This paper is structured as follows.
In Section~\ref{sec:rel_work} we discuss the definition of the label dependency in the multi-label concept and the relevant related work and , and in Section~\ref{sec:method} we introduce autoregressive and non-autoregressive formulations for sequence learning, as well as our non-autoregressive latent-variable model.
In Section~\ref{sec:datasets} we introduce the datasets used in our experiments, and in Sections~\ref{sec:exp} and~\ref{sec:results} we discuss our experimental settings and main findings, respectively.
Finally, in Section~\ref{sec:con} we summarize our main findings and provide avenues for future research.

\section{Related Work}\label{sec:rel_work}
In the probabilistic view, one of the key differences of multi-label learning and binary or multi-class classification is the possibility of occurrence of several labels together.
In other words, in binary and multi-class classification the aim is to only find dependencies between input and output spaces.
However, in multi-label learning the probability of occurrence of one label may impact the probability of another label.
For example, let us say an image has label ``Oscar winner'', and ``South Korea''; that makes it more likely that the label ``Parasite'' is relevant for this image too.
Another example is a video clip labelled ``politics'', which would likely make the probability of the label ``entertainment'' to be lower than usual. Therefore, extracting label dependencies has a crucial role in multi-label learning.

Relevant related work can be divided in three groups based on the dependencies considered between labels: \textit{none}, \textit{pairwise}, and \textit{higher-order label dependency}.
The most general formula for dependency training, i.e. which models all possible interactions between labels, can be written as below.
\begin{eqnarray}\label{eq:baseline_prob}
 P(\bold{y};\theta)=\Pi_{i=1}^{l}P(y_i;\theta|y_{1 : i-1},y_{i+1 : l}).
\end{eqnarray}

As this equation is very expensive, in the following we discuss alternatives to it proposed in the literature. 

\subsection{No Label Dependency}
The simplest solution is to have conditionally independent label probabilities.
\begin{equation}\label{eqn:label_independent}
P_{ind}(\textbf{y};\theta)=\Pi_{i=1}^l P(y_i;\theta).
\end{equation}

In this case, multi-label learning is done label by label without considering label dependencies. The problem is converted into a set of independent binary classification problems and marginal probabilities for each label can be computed efficiently. The most important property of this approach is its simplicity, however label prediction quality will not be very strong as a result of ignoring label dependencies.

\citet{zhang2012composite} suggest to use probabilistic decomposition in order to improve the quality and cost of learning. Composite likelihood is a partial representation of the complete probability function which can be shown as multiplying some components of the marginal or conditional dependencies.
\citet{zhang2012composite} claim that some of the popular multi-label learning approaches can be written using composite likelihoods.
For instance, the binary relevance approach \cite{zhang2010multi} can be modeled as a simple extension of Equation \ref{eqn:label_independent} to include input features as below. 
\begin{equation}\label{eq:BR}
P_{br}(\textbf{y};\theta|\textbf{x})=\Pi_{i=1}^lf(y_i;\theta|\textbf{x}).
\end{equation}

\subsection{Pairwise Label Dependency}

A way to start to address the label dependency issue is to use
pairwise probabilities as follows.
\begin{equation}
P_{pair}(\textbf{y};\theta)=\Pi_{i=1}^{l-1}\Pi_{j=i+1}^l P(y_i,y_j;\theta).
\end{equation}
and conditional pairwise probabilities as below.
\begin{equation}
P_{pcl}(\textbf{y};\theta)=\Pi_{i=1}^{l}\Pi_{j=1}^lf(y_i|y_j;\theta).
\end{equation}

\citet{furnkranz2008multilabel} compute the pairwise ranking of relevant and irrelevant labels by comparing each label to another dummy label, and \citet{hullermeier2008label} models the interaction between each pair of labels directly.
Although learning these label pair correlations can lead to improved results, some times labels have high-order dependencies that cannot be captured with simple pairwise probabilities.

A ranking based approach to pairwise learning can be done as below.
\begin{equation}
P_{pair}(\textbf{y};\theta)=\Pi_{i=1}^{l-1}\Pi_{j=i+1}^l f(y_i\geq y_j|\textbf{x};\theta_{ij}).
\end{equation}

Calibrated pairwise ranking is a combination of one-vs-one and one-vs-all classifiers and can be formulated as follows.
\begin{align}
P_{clr}(\textbf{y};\theta) = &[\Pi_{i=1}^lf(y_i|\textbf{x};\theta_i)] \cdot \notag\\
&[\Pi_{i=1}^{l-1}\Pi_{j=i+1}^l f(y_i\geq y_j|\textbf{x};\theta_{ij})].
\end{align}

Since the parameters in each of these components are independent of each other, we can use maximum likelihood in order to estimate each component independently.


\subsection{Higher Order Dependencies}

The size of the output space, i.e. the number of labels, is the most important challenge in multi-label learning problems since the number of possible outputs grows exponentially with the label space dimension, i.e. there are $2^{l}$ possible interactions between $l$ possible labels.
For example, with $l=20$ possible labels there are $(2^{20})$ possible output patterns.
In order to overcome this complexity, it is necessary to constrain label dependencies to make the problem tractable \cite{zhang2010multi,tsoumakas2010mining}. 
It is clear that higher order approaches are more powerful than first and second order approaches by modelling deeper label dependencies, however they suffer from higher computational cost and lower scalability.

\citet{dembszynski2010label} propose a probabilistic classifier chain (PCC) which decomposes the joint probability of the labels into $L$ prediction tasks which each label is been strengthen by feeding the previous predicted labels by considering a priori given chain order for the labels. 
\begin{eqnarray}\label{eq:chain}
     \log p(\bold{y}|\bold{x}) = \Sigma_{i=1}^{l} \log p(y_i|y_{<i}, \bold{x}),
\end{eqnarray}
\noindent
where $y_{<i}$ denotes the target labels preceding $y_i$.
A na\"{i}ve approach requires $L^2$ path evaluations in the inference step, and is therefore also intractable. However, some approaches have been proposed to reduce the computational complexity \cite{mena2015using}.
Apart from the computational issue, PCC has also a few fundamental problems, one of which is the cascading of errors as the length of the chain gets longer \cite{senge2014problem}.
During training, classifiers in the chain are trained to enrich the input vector $\bold{x}$ with the corresponding true previous targets as additional features.
In contrast, at test time the classifier uses the previously predicted labels to enrich $\mathbf{x}$, which are not guaranteed to be true.
Another key limitation of PCCs is that in practice the label order in a chain has an impact on estimating the conditional probabilities. This issue was addressed in the past by ensemble averaging \cite{dembszynski2010label,read2011classifier} or by a previous analysis of the label dependencies, e.g., by Bayes nets \cite{sucar2014multi}, and selecting the ordering accordingly.
\citet{kumar2013beam} order the chain according to the difficulty of the single-label problems, and \citet{liu2015optimality} use dynamic programming to find the globally optimal label order.

Many of the early proposed multi-label learning approaches struggle with large-scale applications, and either model labels independently of one another or model label dependencies in a way that leads to a costly and complicated model~\cite{tsoumakas2010mining}.
There are several different groups of approaches aiming to utilize high-order label dependencies considering the curse of the label dimensionality, we will elaborate on in the following paragraphs.


\paragraph{Transformation Based}
Transformation-based models propose to project labels into a compressed space, work on that compressed space,
and map results predicted for the smaller space back to the original space.
\citet{hsu2009multi} present the first approach targeting label space compression based on compressed sensing, which assumes sparsity of the label space. An expensive optimisation problem has to be solved in the prediction step. 
\citet{rai2015large} propose probabilistic models to consider label correlation and missing labels in a joint structure.
\citet{tai2012multilabel,chen2012feature,yu2014large,lin2014multi} used orthogonal projections and low-rank assumptions to extract a label matrix decomposition and find a low-dimensional embedding space.
\citet{bhatia2015sparse} performs a local embedding of the label vectors.
To achieve stronger locality, they cluster the data into smaller regions, which is unstable and costly for high-dimensional spaces and one needs an ensemble of the learners to achieve a good prediction accuracy.

Although these approaches make the embedding space smaller and more tractable, they may lead to loss of information as a result of linear transforming the label space to lower-dimensional spaces.
Many of these approaches rely on low-rank assumptions which transform the sparse label space to a new dense embedding space resulting in even lower accuracy, with a higher prediction cost in the new complicated space \cite{bhatia2015locally}.

\paragraph{Structure sparsity} \citet{balasubramanian2012landmark} and \citet{bi2013efficient} proposed to select a subset of the labels, and solve the problem in the original label space, based on structure sparsity optimization and SVD decomposition.
However, these methods are not tractable for large scale data and are not compatible with real application data.
Moreover, training error in the label selection step is ignored, which can lead to the selection of hard-to-predict labels and resulting in training error propagation into the next steps. \citet{barezi2019submodular} proposed greedy label sampling, and \citet{yen2016pd}, instead of making a structural assumption on the relations between labels, assume the label space is highly sparse and has a strong correlation with the feature space, and ignores label space correlations.
\citet{yen2017ppdsparse} proposed the parallel version of \citet{yen2016pd}.

\paragraph{Data partitioning}
Another recent research trend includes methods that partition the data into smaller groups.
\citet{barezi2017multi}, divide the label space into smaller independent groups, while
\citet{agrawal2013multi,prabhu2014fastxml,prabhu2018parabel} propose to partition the data into tree-structured hierarchical groups. Although these partitioning-based approaches avoid information loss, finding a partitioning tree requires solving a complicated optimization problem, which is expensive and needs many training samples. 

In our work, we aim to compare the decomposition methods considering high-order label dependencies together with the curse of label dimensionality. We propose a probabilistic latent-variable based model to utilize high-order label dependencies, and then provide a comprehensive comparison over the large-scale multi-label learning using high-order autoregressive (chain-ordered) decomposition methods vs high-order non-autoregressive (no order is considered for the labels) methods.

\section{Methodology}\label{sec:method}
\subsection{Autoregressive seq2seq learning}
Autoregressive or chain-based label prediction methods belong to a family of encoder–decoder models and encode input features into a fixed-length vector $\bold{x}$ from which a decoder generates a sequence of labels $\bold{y}$,
as described in Equation \ref{eq:chain}.
The model is jointly trained to maximize the probability of the labels' sequence, or only the correct labels, given a source feature vector.\footnote{Predicting only correct labels is more efficient since it requires less prediction steps and does not suffer as much from error propagation at inference time compared to predicting all labels.}
Here, the probability of emitting each label $p(y_i|y_{<i}, \bold{x})$ is parameterized with a recurrent neural network.
To perform inference with this model, one could predict target labels sequentially by greedily taking the \textit{argmax} over the next label prediction probabilities.
Decoding ends when a dummy ``end-of-sequence'' label is predicted. In practice, however, this greedy approach yields sub-optimal results and beam search is often used. However, decoding with a large beam size significantly decreases prediction speed.

\begin{figure}[!htb]
\centering
\includegraphics[trim=230 150 150 210,clip,scale=0.5]{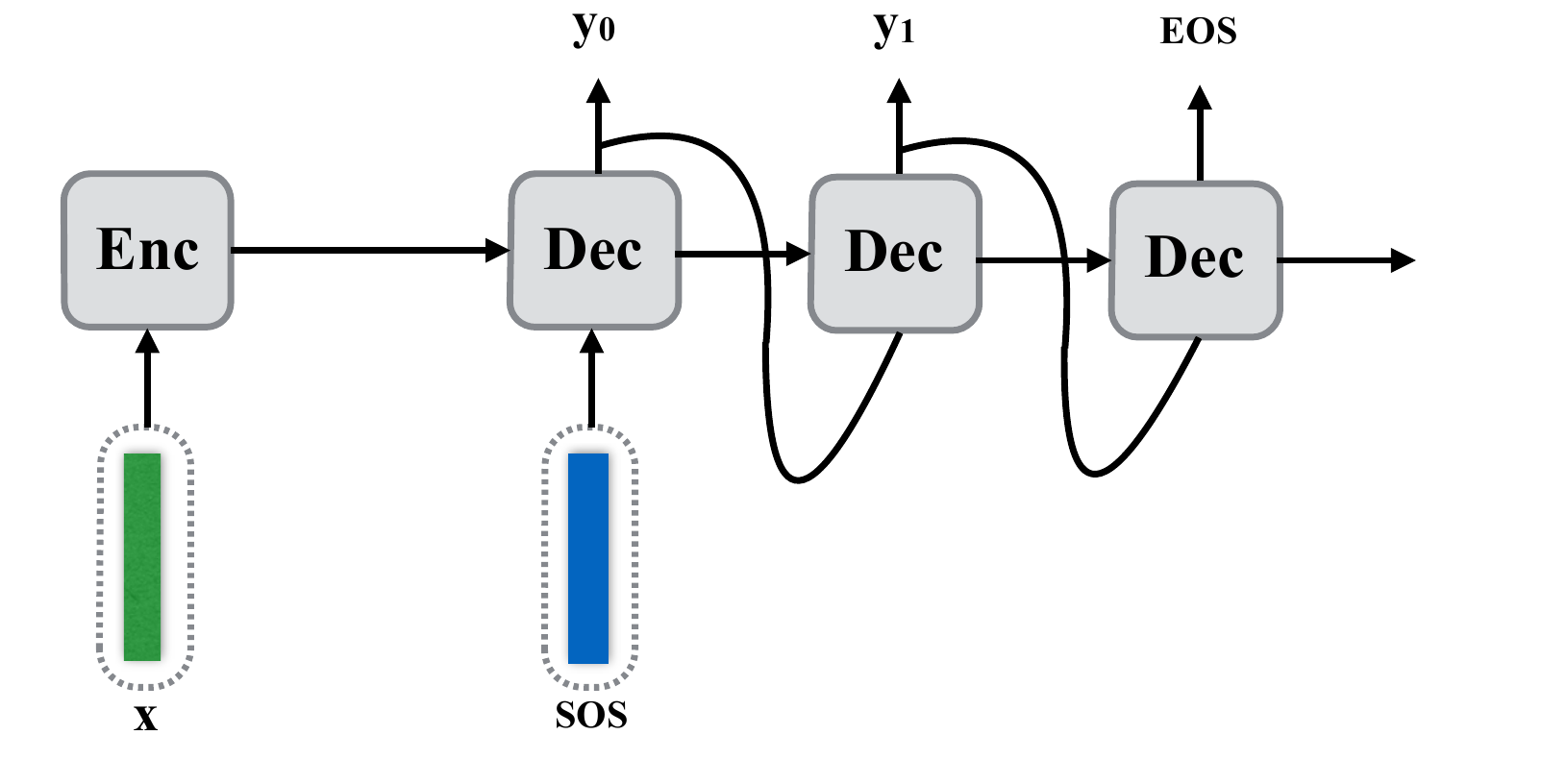}
\caption{Sequence-to-sequence label prediction.}\label{fig:RNN}
\end{figure}

We use a baseline seq2seq model as shown in Figure \ref{fig:RNN}. We encode feature vector $\mathbf{x}$ using a linear encoder and then use an RNN to predict the labels based on a given order.
We have used the default order of the labels as the given order, however according to \citet{vinyals2015order} different orders can lead to different results.

\subsection{Non-autoregressive seq2seq learning}
Although autoregressive models are empirically strong for seq2seq learning, as demonstrated in recent advances in neural machine translation \cite[NMT;][]{wang2016cnn,gehring2017convolutional,nam2017maximizing}, the main drawback is that an autoregressive decoder cannot be used to generate labels in parallel, which results in inefficient use of computational resources and increased latency.
In contrast, non-autoregressive models \cite{wang2016cnn,gehring2017convolutional,nam2017maximizing} predict target tokens without depending on preceding tokens (Equation \ref{eq:BR}).

As the prediction of each target token $y_i$ now depends only on the input features $\bold{x}$, the translation process can be easily parallelized.
We obtain a target sequence simply by taking the \textit{argmax} over all token probabilities, possibly in parallel.
The main challenge of non-autoregressive models is (1) capturing dependencies among target tokens (2) devising strategies to ensure the consistency of the predicted labels.

\subsection{Non-autoregressive latent variable model}
\begin{figure}[!htb]
\centering
\includegraphics[trim=130 600 250 110,clip,scale=0.9]{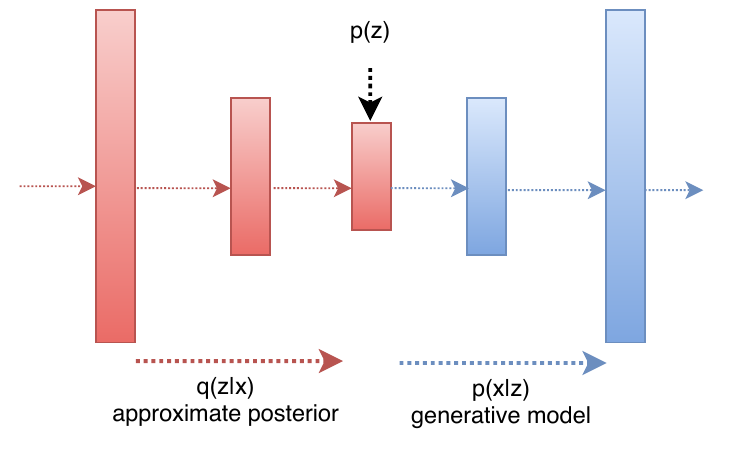}
\caption{A latent-variable model captures higher-order label dependencies. The dependency $p(x|z)$ (blue) of the sequence $x$ on the latent variable $z$ is modeled by a neural network. Inference and learning are made tractable by joint training with an approximate inference network $q(z|x)$(pink). This combination of generative model and inference is known as the variational auto-encoder.}\label{fig:latent}
\end{figure}

We introduce a latent-variable model $p(x,z|\theta)$ to capture higher-order interactions between positions in a sequence.
We imagine that when the data are generated, a hidden variable $z$ is sampled from a prior distribution $p(z)$, in our case a standard multivariate normal, and a sequence $x$ is then generated on the basis of a conditional distribution $p(x|z,\theta)$ that is parameterized by a neural network.
If the system were fully observed, the probability of the data would be simple to compute as $p(x|z,\theta)p(z)$, but when $z$ is hidden we must contend with the marginal likelihood,
$p( {{\mathbf{x}}| {\boldsymbol{\theta }} } ) = {\int} {p( {{\mathbf{x}}| {{\mathbf{z}},{\boldsymbol{\theta }}} } )p({\mathbf{z}})d{\mathbf{z}}}$
which considers all possible explanations for the hidden variables $z$ by integrating them out. Direct computation of this probability is intractable in the general case, but we can use variational inference to compute a lower bound on the (log) probability.
This bound, known as the evidence lower bound (ELBO) ${\cal L}\left( {\boldsymbol{\phi} ;{\mathbf{x}}} \right)$, takes the form
\begin{align}
\log p(\mathbf{x}|\boldsymbol{\theta }) \ge {\cal L}( {\boldsymbol{\phi} ;{\mathbf{x}}})  \mathop{=} \limits^{ \mathrm{\Delta} } & \E_q[ {\log p( {{\mathbf{x}}|{\mathbf{z}},{\boldsymbol{\theta }}} )} ] \notag\\
& - \kl{q ( {{\mathbf{z}}|{\mathbf{x}},\boldsymbol{\phi} } )}{p( {\mathbf{z}} )},
\end{align}
\noindent
where $q(z|x,\phi)$ is a variational approximation for the posterior distribution $p(z|x,\theta)$ of hidden variables given the observed variables. We model both the conditional distribution $p(x|z,\theta)$ of the generative model and the approximate posterior $q(z|x,\phi)$ with neural networks, which results in a flexible model-inference combination known as a variational auto-encoder \cite[VAE;][]{kingma2013auto} as in Fig.~\ref{fig:latent}.

Output labels are assumed to be generated by transforming stochastic latent embeddings $z$.
We propose a non-autoregressive latent-variable model by introducing a sequence of continuous latent variables to model the uncertainty about the target sentence. 
These latent variables $z$ are constrained to have the labels length plus one to account for the input feature vector $\mathbf{x}$, that is, $|z| = |y|+1$.
As there are no observations for $z$, we cannot estimate it directly. We must instead marginalise $z$ out, which yields the following marginal function.
\begin{eqnarray}\label{LL}
P_{nAR}(\textbf{y} | \textbf{x}) &=\int p(\textbf{y},\textbf{z} | \textbf{x}) d\textbf{z} = \int p(\textbf{y} | \textbf{z},\textbf{x}) p(\textbf{z} | \textbf{x}) d\textbf{z} \nonumber \\ 
&= \Pi _{i=1}^{l} \int  p(y_i | \textbf{z},\textbf{x}) p(\textbf{z}| \textbf{x}) d\textbf{z} \nonumber
\end{eqnarray}

\begin{eqnarray}
log P(y|x) & = log \int p(y,z|x) dz \nonumber \\
&= log \int p(y,z|x) \frac{q(z|x,y)}{q(z|x,y)} dz \nonumber \\
&= log(E_{q(z|x,y)} \frac{p(y,z|x)}{q(z|x,y)}) \nonumber \\
& \geq E_{q(z|x,y)} (log(p(y,z|x))-log(q(z|x,y))) \nonumber \\
&= E_{q(z|x,y)} (log(p(y|z,x)p(z|x))-log(q(z|x,y))) \nonumber \\
&= E_{q(z|x,y)} log(p(y|z,x) -KL(q(z|x,y)\|p(z|x)) \nonumber
\end{eqnarray}

Parameter estimation for our model is challenging due to the intractability of the marginal likelihood as a result of continuous latent space.
Instead of directly maximizing the objective function in Equation \ref{LL}, we employ variational inference \cite[VI;][]{jordan1999introduction}, in particular amortised VI \cite{kingma2013auto}, and estimate parameters that maximise a lowerbound on the marginal log-probability function as below.
\begin{eqnarray}\label{ELBO}
 & L(\omega,\phi,\theta) =& 
 \mathbb{E}_{z\sim q_{\phi}}[\log p_{\theta} (y|x,z)] \\
 && - \kl{q_{\phi}(z|x,y)}{p_{\omega}(z|x)},  \nonumber
\end{eqnarray}

\noindent
where $p_{\omega}(z|x)$ is the prior, $q_{\phi}(z|x, y)$ is the approximate posterior and $p_{\theta} (y|x, z)$ is the decoder.
The ELBO objective function in Eq.~\ref{ELBO} is parameterized by three sets of parameters: $\omega$, $\phi$ and $\theta$.
Both the prior $p_{\omega}$ and the approximate posterior $q_{\phi}$ are modeled as spherical Gaussian distributions:
\begin{eqnarray}\label{repp}
 p_{\omega}(z | x) &=& \mathcal{N}(z | \mu_x, \text{diag}(\sigma_x^2)), \\
 \mu_x &=& f_{\mu_x}(x), \nonumber\\
 \sigma_x &=& f_{\sigma_x}(x). \nonumber\\
 q_{\phi}(z | x,y) &=& \mathcal{N}(z | \mu_{xy}, \text{diag}(\sigma_{xy}^2)), \\
 \mu_{xy} &=& g_{\mu_{x,y}}(x,y), \nonumber \\
 \sigma_{xy} &=& g_{\sigma_{x,y}}(x,y), \nonumber
\end{eqnarray}
\noindent
where $\mu_x, \; \sigma_x, \; \mu_{xy}, \;  \sigma_{xy} \in \mathbb{R}^{c}$.
Location-scale variables (e.g. Gaussians) can be reparametrised, i.e. we can obtain a latent sample via a deterministic transformation of the variational parameters and a sample from the standard Gaussian distribution:
\begin{eqnarray}\label{rep2}
z = \mu + \epsilon \sigma^2, \quad \text{where} \quad \epsilon \sim \mathcal{N}(0,I).
\end{eqnarray}

This reparameterization \textit{trick} \cite{kingma2013auto} enables backpropagation through stochastic units so that the model can be trained end-to-end. 

Decoding the output label set is done conditioned on $z$ and input feature vector $x$ by drawing each target label in context from a Categorical observation model:
\begin{eqnarray}\label{dec}
    p_{\theta}(y|x,z) &=& \text{Cat}(\pi),\\
    \pi &=& f_{\pi}(x,z).\nonumber
\end{eqnarray}

\subsubsection{A Modified Objective Function with Length Prediction}
During training, we want the model to maximize the lowerbound in Eq.~\ref{ELBO}.
However, to generate a label set, the target length $l_y$ has to be predicted first.
We let the latent variables model the target length by parameterizing the decoder as:
\begin{eqnarray}\label{Pl}
p_{\theta}(y|x, z) &=& \Sigma_{l_y} p_{\theta}(y, l_y|x, z), \label{Pl1}\\
p_{\theta}(y, l_y|x, z) &=& p_{\theta}(y|x,z,l_y) p_{\theta}(l_y|z) \label{Pl2}
\end{eqnarray}
Here $l_y$ denotes the length of $y$.
This step helps to predict the length of the output sequence which previously had been done by simply binarizing predictions by using a threshold for the outputs, or by adding an extra dummy label as a threshold for positive and negative labels \cite{furnkranz2008multilabel}.
Equation \ref{Pl1} is valid as the probability $p_{\theta}(l \neq l_y|x, z)$ is always zero. Plugging in Eq.~\ref{Pl} with the independence assumption on both latent variables and target tokens, the objective has the following form:
\begin{eqnarray}\label{ELBOl}
     L(\omega,\phi,\theta) = &\E_{z\sim q_{\phi}}[\Sigma_{i=1}^{|y|}\log p_{\theta} (y_i|x,z,l_y)+ \notag\\
     & \log p_{\theta}(l_y|z)] \\
     & - \kl{q_{\phi}(z|x,y)}{p_{\omega}(z|x)}. \nonumber
\end{eqnarray}

\subsection{Transformers and Self-Attention}
In this section we discuss the transformer architecture, i.e. a model that uses stacks of self-attention layers and which naturally lends themselves to parallelization.

A transformer encoder operates on a sequence of vectors that in our task would be the label embeddings concatenated to the input feature vector. We compute self-attention over a concatenation of the $|y|$ label representations and the input feature vector representations $x$, which results in a matrix of $|y|+1$ representation vectors. 
The transformer layer has two main components: a self-attention and a feed-forward layer.

\paragraph{Self-attention layer}
Transformer-based architectures, which are primarily used in modelling language understanding tasks, eschew the use of recurrence in neural network and instead trust entirely on self-attention mechanisms to draw global dependencies between inputs and outputs. Their greatest benefit comes from how transformers lends themselves to parallelization.
By concatenating labels and input feature vector and using that as input to a transformer layer in a multi-label learning task, we can efficiently extract and benefit from the label-label and label-feature dependencies.
Self-attention is the method the transformer uses to bake the ``understanding'' of other relevant vectors into the one that currently is processing as is shown in Equation \ref{eq:self_att}, where K,V and Q are \textit{key}, \textit{value} and \textit{query}, respectively, and $n=|y|+1$ is the input sequence length.
\begin{equation}\label{eq:self_att}
    \text{Attention}(\mathbf{Q}, \mathbf{K}, \mathbf{V}) = \text{softmax}(\frac{\mathbf{Q}\mathbf{K}^\top}{\sqrt{n}})\mathbf{V}
\end{equation}

We use transformers to employ self-attention over both input and label spaces simultaneously. We need to score each position in the above explained $|y|+1$ representation vector against all other positions. The score determines how much focus to place on different parts of the input vector (labels and input features) as we encode each label.
The final score determines how much each part of the input will impact the current label.

\paragraph{Feed-forward layer}
The outputs of the self-attention layer are fed to a feed-forward neural network.
The same feed-forward network is independently applied to each position, then sends out the output upwards to the next unit.

\paragraph{Positional embedding} Since in our multi-label classification task there is no order information, we do not need to use positional embeddings
\citep{vaswani2017attention}.
Moreover, since the nature of our labels are set-like not sequential, we do not need to worry about masking future information and directly use unmasked bi-directional information propagation. 

\subsection{Model Architecture}
\begin{figure}[!htb]
\centering
\includegraphics[trim=230 40 230 150,clip,scale = 1, width=.5\textwidth]{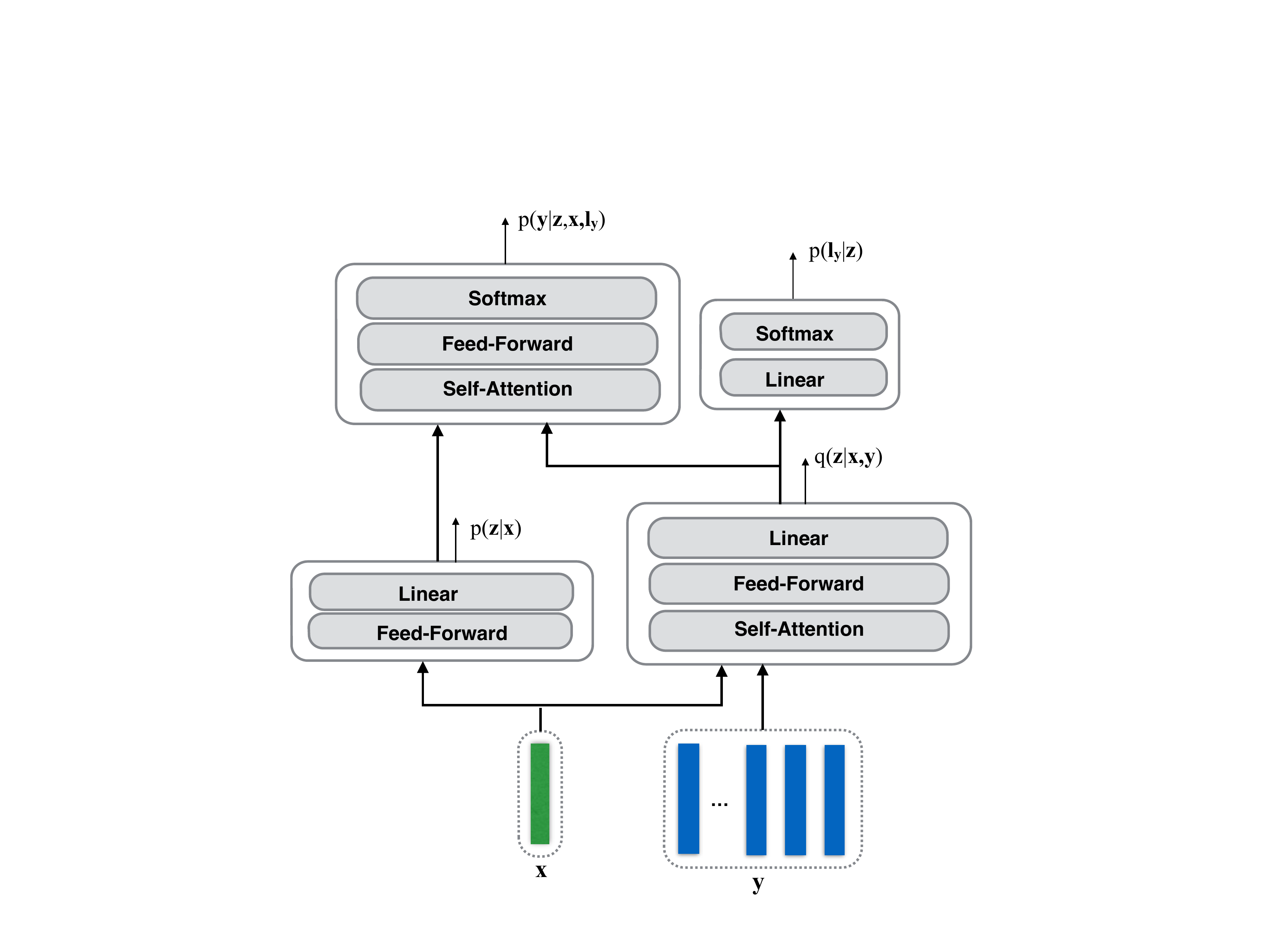}
\caption{Architecture of the non-auto-regressive model}\label{VAE_model}
\end{figure}

As evident from Eq.~\ref{ELBOl}, there are four parameterized components in our model: prior $p_{\omega}(z|x)$, approximate posterior $q_{\phi}(z|x, y)$, decoder $p_{\theta}(y|x, z, l_y)$ and length predictor $p_{\theta}(l_y|z)$.
The architecture of the proposed non-autoregressive model is depicted in Fig.~\ref{VAE_model}, which reuses the encoder part in transformer \cite{vaswani2017attention} to compute the aforementioned distributions.
All of our parametric functions are neural network architectures.
Gaussian layers $f_{\mu_x},f_{\sigma_x},g_{\mu_{x,y}},g_{\sigma_{x,y}}$ are feed-forward networks with a single ReLU hidden layer. 
We use a self-attention architecture based on the encoder part of the transformer models in order to encode the concatenation of the input features $\bold{x}$ and the output labels $\bold{y}$ which leads to a set of size $|\bold{y}|+1$.

For the generative model, $f_{\mu_x}$ and $f_{\sigma_x}$ transform the average input feature's encoder hidden state. We use the self-attention architecture in order to also condition on the label set. Then $g_{\mu_{x,y}}$ and $g_{\sigma_{x,y}}$ transform a concatenation of the average input feature encoder hidden state, and the average label set transformer encoder hidden state.

In the inference step, we use a deterministic iterative algorithm, similar to \citet{shu2020latent}, to refine the approximate posterior over the latent variables and obtain better target predictions. At the inference step, we first do the regular prediction by obtaining the initial posterior from a prior distribution $p(z|x)$ and the initial guess of the target output from $p(y|x, z)$. Then for refining the output, we alternate between updating the approximate posterior and target output with the help of the approximate posterior $q(z|x,y)$. The inference method is deterministic as we use $z=\mu$. This iterative methods significantly improves the output by providing the opportunity to refine the latent variables which leads to refining both the length of the output and output labels. 

\section{Datasets}\label{sec:datasets}
We used five different datasets in our experiments.
\textbf{Bibtex} is a text dataset extracted from the BibSonomy website \cite{katakis2008multilabel} and contains metadata for bibtex items (e.g. title of the paper, authors, etc.) and provides term-frequency features. \textbf{Mediamill} is derived from the Mediamill contest datasets, which include low-level multimedia features (visual and textual features) extracted from 85 hours of international news videos from the TRECVID 2005/2006 benchmark datasets \cite{snoek2006challenge} labeled using 101 semantic concepts, like commercials, nature, and baseball. 
\textbf{Eurlex} includes 19348 legal documents from European nations, containing several different types, including treaties, legislation, case-law and legislative proposals, classified according to the EUROVOC descriptor using 3993 different classes, and 5000 features extracted using common TF-IDF term weighting \cite{mencia2008efficient}.
\textbf{Delicious} is a text dataset extracted from the {\tt del.icio.us}  social bookmarking site on the 1st of April 2007 and contains textual data of web pages along with their user defined tags \cite{tsoumakas2008effective}. The content of web pages was represented using the Boolean bag-of-words model.
Finally, \textbf{Wiki10-31K} is a collection of social tags for given Wikipedia pages with TF-IDF features \cite{zubiaga2012enhancing}.
Datasets statistics are provided in Table \ref{tbl_data}.

\begin{table}[!tb]
\centering
\resizebox{0.47\textwidth}{!}{
\begin{tabular}{ cccccc } 
\toprule
\multirow{2}{*}{\textbf{Dataset}}	  & \multirow{2}{*}{\textbf{Domain}}	& \multirow{2}{*}{\textbf{\# Features}}   &	\multirow{2}{*}{\textbf{\# Labels}}  &    \textbf{\# Training}  & \textbf{\# Testing}	\\
 & 	&   &	&    \textbf{examples}  & \textbf{examples}	\\
\midrule
Bibtex	  & Text	& 1836	 	&  159	   &    4880	      &     2515	\\
Delicious &	Text(Web)& 500	    &  983	   &    12920	      &     3185	\\
Mediamill &	Video 	& 120		&	101	   &    30993		  &	    12914   \\
Eurlex    & Text	& 5000     	&  3993    &    17413         &     1935    \\
\midrule
Wiki10-31K  & Text	& 101938	&  30938   &	14146		  &	    6616    \\
\bottomrule
\end{tabular}}\caption{Datasets statistics.}\label{tbl_data}
\end{table}

\section{Experiments}\label{sec:exp}
\subsection{Experimental Setup}
For the small datasets (Bibtex, Mediamill, Delicious and Eurlex), we report the average of 10 different experiments for random partitions of each dataset.
For the larger dataset Wiki10-31K, we report one experiment with the training and testing partition as in Table~\ref{tbl_data}. 
All network's hyperparameters are chosen according to validation set precision by using 10-fold cross validation, and using a greedy search.

\begin{table*}[!htb]
\scriptsize 
\centering
\begin{tabular}{ lccccccccccc }
\toprule
&& \textbf{Seq2Seq} & \textbf{nAR} & \textbf{SLEEC} &	\textbf{FastXML} & \textbf{PD-sparse} &	\textbf{LEML} &	\textbf{CPLST} &	\textbf{CS} &	\textbf{ML-CSSP} \\
\midrule

\multirow{3}{*}{\textbf{Bibtex}} & P@1	&  \textbf{93.57}$\pm$0.73 & 49.92$\pm$1.3 & 65.08$\pm$0.65 &  63.42$\pm$0.67 &  61.29$\pm$0.65  & 62.54$\pm$0.52 & 62.38$\pm$0.63 & 58.87$\pm$0.61 & 44.98$\pm$1.15   \\
& P@3 & 36.65$\pm$0.95 & 28.49$\pm$0.73 & \textbf{39.64}$\pm$0.39 &  39.23$\pm$0.57 & 35.82$\pm$0.46  & 38.41$\pm$0.42 & 37.84$\pm$0.48 & 33.53$\pm$0.49 & 30.43$\pm$0.59   \\
& P@5	& 21.99$\pm$0.57 & 20.67$\pm$0.67 & \textbf{28.87}$\pm$0.32&  28.86$\pm$0.38  & 25.74$\pm$0.30 & 28.21$\pm$0.24 & 27.62$\pm$0.27 & 23.72$\pm$0.29 & 23.53$\pm$0.37     \\
\midrule
\multirow{3}{*}{\textbf{Delicious}} & P@1	& \textbf{97.36}$\pm$0.37 & 61.78$\pm$0.91 & 67.59$\pm$0.53 & 69.61$\pm$0.58 & 51.82$\pm$1.40  & 65.67$\pm$0.73 & 65.31$\pm$0.88 & 61.36$\pm$0.38 & 63.04$\pm$1.28  \\
& P@3	& \textbf{69.40}$\pm$0.98 & 56.24$\pm$0.84 & 61.38$\pm$0.59 & 64.12$\pm$0.75  & 44.18$\pm$1.04  & 60.55$\pm$0.48 & 59.95$\pm$0.43 & 56.46$\pm$0.33 & 56.26$\pm$1.13 \\
& P@5	& 45.07$\pm$0.93 & 51.46$\pm$0.84 & 56.56$\pm$0.54 & \textbf{59.27}$\pm$0.65 & 38.95$\pm$0.94  & 56.08$\pm$0.43 & 55.31$\pm$0.50 & 52.07$\pm$0.30 & 50.16$\pm$0.83  \\
\midrule
\multirow{3}{*}{\textbf{Mediamill}} & P@1	& \textbf{93.57}$\pm$0.35 & 77.14$\pm$0.21 & 87.82$\pm$0.33 & 84.22$\pm$0.27 & 81.86$\pm$4.08  & 84.01$\pm$0.31 & 83.35$\pm$0.33 & 83.82$\pm$5.92 & 78.95$\pm$0.23  \\
& P@3	& 71.81$\pm$1.23 & 58.64$\pm$0.32 & \textbf{73.45}$\pm$0.30 & 67.33$\pm$0.20& 62.52$\pm$2.31  & 67.20$\pm$0.23 & 66.18$\pm$0.22 & 67.32$\pm$4.42 & 60.93$\pm$0.24  \\
& P@5	 & 46.55$\pm$0.97 & 46.12$\pm$0.11 &  \textbf{59.17}$\pm$0.34 & 53.04$\pm$0.18 & 45.11$\pm$1.14 & 52.80$\pm$0.18 & 51.46$\pm$0.20 & 52.80$\pm$2.61 & 44.27$\pm$0.20  \\
\midrule
\multirow{3}{*}{\textbf{Eurlex}} & P@1	& \textbf{83.18}$\pm$0.97 & 54.93$\pm$6.06  & 79.26$\pm$0.86 & 71.36$\pm$1.63 & 76.43$\pm$1.04  & 63.40$\pm$1.58 & 72.28$\pm$0.99 & 58.52$\pm$1.06 & 62.09$\pm$2.12 \\
& P@3	& 46.90$\pm$2.89 & 41.50$\pm$4.89 & \textbf{64.30}$\pm$0.88 & 59.90$\pm$1.58 & 60.37$\pm$0.74  & 50.35$\pm$1.44 & 58.16$\pm$1.11 & 45.51$\pm$0.71 & 48.39$\pm$1.31  \\
& P@5	& 28.56$\pm$1.88 & 33.61$\pm$4.14& \textbf{52.33}$\pm$0.80 & 50.39$\pm$1.40 & 49.72$\pm$0.74  & 41.28$\pm$1.07 & 47.73$\pm$0.97 & 32.47$\pm$0.58 & 40.11$\pm$1.10  \\
\midrule
\multirow{3}{*}{\textbf{Wiki10-31k}} & P@1	& 74.16$\pm$2.37 & 76.03 & \textbf{85.88} &  83.03 & 82.14 & 73.47 & - & - & - \\
& P@3	& 27.88$\pm$0.29 & 64.63 & \textbf{72.98} &  67.47 & 69.68 & 62.43 & - & - & - \\
& P@5	& 16.75$\pm$0.18 & 56.60 &\textbf{62.70} &  57.76 & 58.76 & 54.35 & - & - & - \\
\bottomrule
\end{tabular} \caption{Validation set precision@k for single models. Best in \textbf{bold}.}\label{tbl_smallp1}
\end{table*}

\textbf{Baselines.} The proposed autoregressive and non autoregressive models were compared with several state-of-the-art methods with diverse approaches:
\textbf{LEML} \cite{yu2014large}, \textbf{CPLST} \cite{chen2012feature}, \textbf{CS} \cite{hsu2009multi} and \textbf{SLEEC} \cite{bhatia2015sparse}, all of them embedding-based approaches with a low-rank or sparse assumption in the label space;
\textbf{ML-CSSP} \cite{bi2013efficient}, which solves the problem in the original label space and ignores training error in subset selection step;
\textbf{FastXML} \cite{prabhu2014fastxml} and \textbf{PD-sparse} \cite{yen2016pd}, which do not use an embedding transformation and aim to solve the problem without compression or sampling. 
For the baseline methods, we use the reported results as provided by the authors and \cite{bhatia2015locally}.

\textbf{Evaluation metrics.} We use \textit{precision$@k$} (P$@k$) and \textit{nDCG$@k$} as in Eqs. \ref{p@k} and \ref{ndcg@k}, and their \textit{propensity-scored} versions as in Eqs. \ref{psp@k} and \ref{psndcg@k}, to evaluate the results.
\begin{eqnarray}
    \text{P}@k &:= \frac{1}{k} \sum_{l\in \text{rank}_k (\hat{\mathbf y})} \mathbf y_l  \label{p@k}\\
    \text{PSP}@k &:= \frac{1}{k} \sum_{l\in \text{rank}_k (\hat{\mathbf y})} \frac{\mathbf y_l}{p_l} \label{psp@k} \\
\text{DCG}@k &:= \sum_{l\in {\text{rank}}_k (\hat{\mathbf y})} \frac{\mathbf y_l}{\log(l+1)} \nonumber  \\
\text{nDCG}@k &:= \frac{{\text{DCG}}@k}{\sum_{l=1}^{\min(k, \|\mathbf y\|_0)} \frac{1}{\log(l+1)}} \label{ndcg@k} \\
\text{PSDCG}@k &:= \sum_{l\in {\text{rank}}_k (\hat{\mathbf y})} \frac{\mathbf y_l}{p_l\log(l+1)} \nonumber \\
\text{PSnDCG}@k &:= \frac{{\text{PSDCG}}@k}{\sum_{l=1}^{k} \frac{1}{\log(l+1)}} \label{psndcg@k}
\end{eqnarray}

\section{Results and Discussion}\label{sec:results}

In Table~\ref{tbl_smallp1}, we show the average and standard deviation of precision@k scores for the four small-scale datasets (Bibtex, Mediamill, Delicious and Eurlex) and for the large-scale dataset Wiki10-31k. 
For Wiki10-31k, results are reported only for those baselines that were tractable (i.e. PD-sparse and LEML).
Results for nDCG@k are included in Table~\ref{tbl_smalld1} in Appendix.
We highlight that the seq2seq (autoregressive) model consistently outperforms the other state-of-the-art and baseline models, except under two conditions:
  (1) First, the prediction quality for predicting more than 1 label, that is precision@k and nDCG@k for $k>1$.
  (2) Second, the precision@k and nDCG@k for all values of $k$ for the large scale datasets like Wiki10. We can conclude that the seq2seq model is strong in predicting highly ranked labels, however its performance weakens with the increase in the number of labels (tags) for the sample.
We note that SLEEC and FastXML, which are better than the seq2seq model in some cases, are ensembles and methods based on transforming the label space and tree-based partitioning of the labels, correspondingly. Otherwise, the seq2seq model beats the baselines for $k>1$ too.

We present the results for auto-regressive and non-autoregressive models with comparable architectures in Tables
\ref{tab:baseline_seq2seq_precisions_no_propensity} and 
\ref{tab:baseline_seq2seq_ndcg_no_propensity} to provide a better comparison of these two approaches in solving multi-label learning tasks.
In Tables~\ref{tab:baseline_seq2seq_precisions_with_propensity} and \ref{tab:baseline_seq2seq_ndcg_with_propensity}, we report propensity scored precision@k and nDCG@k scores on the validation set for all data sets.

These results corroborate the above results and discussion, as the seq2seq model is always the winner except when the number of the labels gets larger or when we measure the prediction accuracy for a larger number of the labels (i.e., large $k$). This results brings a new question to the field: how can we provide a better interpretation of the architectures and know what makes seq2seq model strong to predict the highest ranked label and what makes the non-autoregressive model's weak in predicting the label with the highest rank while it is strong in predicting the other labels in comapare to the autoregressive models?  

\begin{table}[!htb]
\scriptsize 
\centering
\begin{tabular}{ lccc }
\toprule
&& \textbf{Seq2Seq} & \textbf{nAR} \\
\midrule

\multirow{3}{*}{\textbf{Bibtex}}
& P@1 & \textbf{93.57}$\pm$0.73 & 49.92$\pm$1.3 \\
& P@3 & \textbf{36.65}$\pm$0.95 & 28.49$\pm$0.73 \\
& P@5 & \textbf{21.99}$\pm$0.57 & 20.67$\pm$0.67 \\
\midrule
\multirow{3}{*}{\textbf{Delicious}} 
& P@1	& \textbf{97.36}$\pm$0.37 & 61.78$\pm$0.91 \\
& P@3	& \textbf{69.40}$\pm$0.98 & 56.24$\pm$0.84 \\
& P@5	& 45.07$\pm$0.93 & \textbf{51.46}$\pm$0.84 \\
\midrule
\multirow{3}{*}{\textbf{Mediamill}} 
& P@1	& \textbf{93.57}$\pm$0.35 & 77.14$\pm$0.21 \\
& P@3	& \textbf{71.81}$\pm$1.23 & 58.64$\pm$0.32 \\
& P@5	& \textbf{46.55}$\pm$0.97 & \textbf{46.12}$\pm$0.11 \\
\midrule
\multirow{3}{*}{\textbf{Eurlex}} 
& P@1	& \textbf{83.18}$\pm$0.97 & 54.93$\pm$6.06 \\
& P@3	& \textbf{46.90}$\pm$2.89 & 41.50$\pm$4.89 \\
& P@5	& 28.56$\pm$1.88 & \textbf{33.61}$\pm$4.14 \\
\midrule
\multirow{3}{*}{\textbf{Wiki10-31k}} 
& P@1	& 74.16$\pm$2.37 & \textbf{76.03} \\
& P@3	& 27.88$\pm$0.29 & \textbf{64.63} \\
& P@5	& 16.75$\pm$0.18 & \textbf{56.60} \\
\bottomrule
\end{tabular} \caption{Mean and standard deviations precision@k over 10 different publicly-available train/validation splits.}\label{tab:baseline_seq2seq_precisions_no_propensity}
\end{table}

\begin{table*}[!htb]
\scriptsize 
\centering
\begin{tabular}{ lccccccccccc }
\toprule
&& \textbf{Seq2Seq} & \textbf{nAR} & \textbf{SLEEC} &	\textbf{FastXML} & \textbf{PD-sparse} &	\textbf{LEML} &	\textbf{CPLST} &	\textbf{CS} &	\textbf{ML-CSSP} \\
\midrule

\multirow{3}{*}{\textbf{Bibtex}} & 
nDCG@1	 & \textbf{93.57}$\pm$0.73 & 49.94$\pm$1.33 &  65.08 & 63.42 & 61.29$\pm$0.65  & 62.54$\pm$0.52 & 62.38$\pm$0.63 & 58.87$\pm$0.61 & 44.98$\pm$1.15  \\ 
& nDCG@3   & \textbf{71.58}$\pm$1.28 & 28.49$\pm$1.21 & 60.47 & 59.51 & 55.83$\pm$0.57  & 58.22$\pm$0.42 & 57.63$\pm$0.56 & 52.19$\pm$0.56 & 44.67$\pm$1.01 \\ 
& nDCG@5	 & \textbf{69.12}$\pm$1.22 & 20.67$\pm$1.51& 62.64 & 61.70 & 57.35$\pm$0.49  & 60.53$\pm$0.38 & 59.71$\pm$0.42 & 53.25$\pm$0.54 & 47.97$\pm$0.98  \\ 
\midrule
\multirow{3}{*}{\textbf{Delicious}} & 
nDCG@1	&\textbf{97.36}$\pm$0.37 & 61.78$\pm$0.91& 67.59 & 69.61& 51.82$\pm$1.40  & 65.67$\pm$0.73 & 65.31$\pm$0.88 & 61.36$\pm$0.38 & 63.04$\pm$1.29  \\ 
& nDCG@3	& \textbf{76.23}$\pm$0.69 & 57.56$\pm$0.83& 62.87 & 65.47& 46.00$\pm$1.12  & 61.77$\pm$0.50 & 61.16$\pm$0.45 & 57.66$\pm$0.34 & 57.91$\pm$1.15 \\
& nDCG@5	&57.73$\pm$0.71 & 53.98$\pm$0.82& 59.28 & \textbf{61.90}& 42.02$\pm$1.01  & 58.47$\pm$0.47 & 57.80$\pm$0.49 & 54.44$\pm$0.32 & 53.36$\pm$0.94  \\ 
\midrule
\multirow{3}{*}{\textbf{Mediamill}} &
nDCG@1	 &\textbf{93.57}$\pm$0.35 & 77.14$\pm$0.21&87.82 & 84.22 &  81.86$\pm$4.08  & 84.01$\pm$0.31 & 83.35$\pm$0.33 & 83.82$\pm$5.92 & 78.95$\pm$0.23 	 \\ 
& nDCG@3	 & \textbf{82.81}$\pm$0.79 & 58.64$\pm$0.69& 81.50 & 75.41 &  70.21$\pm$2.37  & 75.23$\pm$0.25 & 74.21$\pm$0.24 & 75.29$\pm$4.99 & 68.97$\pm$0.28  \\ 
& nDCG@5	 &  71.88$\pm$0.75 & 46.12$\pm$0.23 &\textbf{79.22} & 72.37 &  63.71$\pm$1.73  & 71.96$\pm$0.18 & 70.55$\pm$0.17 & 71.92$\pm$4.03 & 62.88$\pm$0.26  \\ 
\midrule
\multirow{3}{*}{\textbf{Eurlex}} &
nDCG@1	 & \textbf{83.18}$\pm$0.97 & 54.92$\pm$6.06&79.26 & 71.36 &  76.43$\pm$1.04  & 63.40$\pm$1.58 & 72.28$\pm$0.99 & 58.52$\pm$1.06 & 62.09$\pm$2.12  \\ 
& nDCG@3	 & 55.61$\pm$2.32 & 44.69$\pm$5.14& \textbf{68.13} & 62.87  &  64.31$\pm$0.72  & 53.56$\pm$1.47 & 61.64$\pm$1.02 & 48.67$\pm$0.75 & 51.63$\pm$1.31  \\ 
& nDCG@5	 & 42.67$\pm$1.86 & 40.06$\pm$4.75&\textbf{61.60} & 58.06 &  58.78$\pm$0.70  & 48.47$\pm$1.24 & 55.92$\pm$0.97 & 40.79$\pm$0.65 & 47.11$\pm$1.10  \\ 
\midrule
\multirow{3}{*}{\textbf{Wiki10-31k}} &
nDCG@1	&  74.16$\pm$2.37 & 76.03& \textbf{85.88} & 84.31 & 82.14  & 73.47 & - & - & -   \\ 
& nDCG@3	& 37.56$\pm$0.36 & 67.22& 72.98 & \textbf{75.35}  & 72.63  & 64.92 & - & - & -   \\ 
& nDCG@5	& 27.17$\pm$0.25 & 61.00& 62.70 & 63.36 & \textbf{64.33}  & 58.69 & - & - & -   \\
\bottomrule
\end{tabular} \caption{nDCG@k on the small-scale datasets. Best in \textbf{bold}.}\label{tbl_smalld1}
\end{table*} 
\begin{table}[!htb]
\scriptsize 
\centering
\begin{tabular}{ lccc }
\toprule
&& \textbf{Seq2Seq} & \textbf{nAR} \\
\midrule

\multirow{3}{*}{\textbf{Bibtex}}
& nDCG@1    & \textbf{93.57}$\pm$0.73 & 49.94$\pm$1.33 \\
& nDCG@3    & \textbf{71.58}$\pm$1.28 & 28.49$\pm$1.21 \\
& nDCG@5    & \textbf{69.12}$\pm$1.22 & 20.67$\pm$1.51 \\
\midrule
\multirow{3}{*}{\textbf{Delicious}} 
& nDCG@1    & \textbf{97.36}$\pm$0.37 & 61.78$\pm$0.91 \\
& nDCG@3    & \textbf{76.23}$\pm$0.69 & 57.56$\pm$0.83 \\
& nDCG@5    & \textbf{57.73}$\pm$0.71 & 53.98$\pm$0.82 \\
\midrule
\multirow{3}{*}{\textbf{Mediamill}} 
& nDCG@1	& \textbf{93.57}$\pm$0.35 & 77.14$\pm$0.21 \\
& nDCG@3	& \textbf{82.81}$\pm$0.79 & 66.05$\pm$0.69 \\
& nDCG@5	& \textbf{71.88}$\pm$0.75 & 63.51$\pm$0.23 \\
\midrule
\multirow{3}{*}{\textbf{Eurlex}} 
& nDCG@1	& \textbf{83.18}$\pm$0.97 & 54.92$\pm$6.06 \\
& nDCG@3	& \textbf{55.61}$\pm$2.32 & 44.69$\pm$5.14 \\
& nDCG@5	& \textbf{42.67}$\pm$1.86 & 40.06$\pm$4.75 \\
\midrule
\multirow{3}{*}{\textbf{Wiki10-31k}} 
& nDCG@1	& 74.16$\pm$2.37 & \textbf{76.03} \\
& nDCG@3	& 37.56$\pm$0.36 & \textbf{67.22} \\
& nDCG@5	& 27.17$\pm$0.25 & \textbf{61.00} \\
\bottomrule
\end{tabular} \caption{Single model nDCG@k. Showing mean and standard deviations over 10 different publicly-available train/validation splits. Best in \textbf{bold}.}\label{tab:baseline_seq2seq_ndcg_no_propensity}
\end{table}

\begin{table}[!htb]
\scriptsize 
\centering
\begin{tabular}{ lccc }
\toprule
&& \textbf{Seq2Seq} & \textbf{nAR} \\
\midrule

\multirow{3}{*}{\textbf{Bibtex}}
& PSP@1 & \textbf{84.41}$\pm$0.69 & 36.42$\pm$1.10 \\
& PSP@3 & \textbf{51.37}$\pm$1.53 & 36.09$\pm$1.21 \\
& PSP@5 & \textbf{45.36}$\pm$1.33 & 39.78$\pm$1.77 \\
\midrule
\multirow{3}{*}{\textbf{Delicious}} 
& PSP@1	& \textbf{62.79}$\pm$0.34 & 23.15$\pm$0.43 \\
& PSP@3	& \textbf{45.63}$\pm$0.55 & 24.48$\pm$0.42 \\
& PSP@5	& \textbf{32.15}$\pm$0.59 & 25.06$\pm$0.46 \\
\midrule
\multirow{3}{*}{\textbf{Mediamill}} 
& PSP@1	& \textbf{86.41}$\pm$0.36 & 55.01$\pm$0.11 \\
& PSP@3	& \textbf{74.97}$\pm$1.25 & 53.12$\pm$0.23 \\
& PSP@5	& \textbf{59.23}$\pm$1.19 & 52.45$\pm$0.09 \\
\midrule
\multirow{3}{*}{\textbf{Eurlex}} 
& PSP@1	& \textbf{34.13}$\pm$0.42 & 27.54$\pm$4.66 \\
& PSP@3	& \textbf{27.22}$\pm$1.96 & 25.97$\pm$4.37 \\
& PSP@5	& 21.03$\pm$1.61 & \textbf{26.17}$\pm$4.40 \\
\midrule
\multirow{3}{*}{\textbf{Wiki10-31k}} 
& PSP@1	& 8.37$\pm$0.25 & \textbf{14.15} \\
& PSP@3	& 3.69$\pm$0.04 & \textbf{13.42} \\
& PSP@5	& 2.55$\pm$0.03 & \textbf{13.34} \\
\bottomrule
\end{tabular} \caption{Single model propensity scored precision@k. Showing mean and standard deviations over 10 different publicly-available train/validation splits. Best in \textbf{bold}}\label{tab:baseline_seq2seq_precisions_with_propensity}
\end{table}

\begin{table}[!htb]
\scriptsize 
\centering
\begin{tabular}{ lccc }
\toprule
&& \textbf{Seq2Seq} & \textbf{nAR} \\
\midrule

\multirow{3}{*}{\textbf{Bibtex}}
& PSnDCG@1	& \textbf{84.41}$\pm$0.69 & 36.42$\pm$1.10 \\
& PSnDCG@3	& \textbf{67.06}$\pm$1.29 & 35.96$\pm$1.15 \\
& PSnDCG@5	& \textbf{65.50}$\pm$1.25 & 38.07$\pm$1.42 \\
\midrule
\multirow{3}{*}{\textbf{Delicious}} 
& PSnDCG@1	& \textbf{62.79}$\pm$0.34 & 23.15$\pm$0.43 \\
& PSnDCG@3	& \textbf{49.93}$\pm$0.36 & 24.48$\pm$0.41 \\
& PSnDCG@5	& \textbf{40.14}$\pm$0.42 & 25.06$\pm$0.44 \\
\midrule
\multirow{3}{*}{\textbf{Mediamill}} 
& PSnDCG@1	& \textbf{86.41}$\pm$0.36 & 55.01$\pm$0.11 \\
& PSnDCG@3	& \textbf{79.15}$\pm$0.72 & 53.47$\pm$0.53 \\
& pSnDCG@5	& \textbf{70.41}$\pm$0.71 & 54.10$\pm$0.15 \\
\midrule
\multirow{3}{*}{\textbf{Eurlex}} 
& PSnDCG@1	& \textbf{34.13}$\pm$0.42 & 27.54$\pm$4.66 \\
& pSnDCG@3	& \textbf{29.47}$\pm$1.51 & 26.35$\pm$4.44 \\
& PSnDCG@5	& 25.74$\pm$1.37 & \textbf{26.42}$\pm$4.65 \\
\midrule
\multirow{3}{*}{\textbf{Wiki10-31k}} 
& PSnDCG@1	& 8.37$\pm$0.25 & \textbf{14.15} \\
& PSnDCG@3	& 4.78$\pm$0.03 & \textbf{13.58} \\
& PSnDCG@5	& 3.83$\pm$0.02 & \textbf{13.49} \\
\bottomrule
\end{tabular} \caption{Single model propensity scored nDCG@k. Showing mean and standard deviations over 10 different publicly-available train/validation splits. Best in \textbf{bold} }\label{tab:baseline_seq2seq_ndcg_with_propensity}
\end{table}

\section{Conclusion and Future Work}\label{sec:con}
In this work, we provide an overview on auto-regressive and non-autoregressive multi-label learning methods.
As Inference with autoregressive seq2seq models cannot be easily parallelized due to the autoregressive factorization of the output probability, and AR models tend to struggle with increased output sequence lengths, We have proposed a variational auto-encoder non-autoregressive model and compared it to a seq2seq auto-regressive model to provide a general comparison framework for order-based and order-free multi-label learning.

Extensive experiments on five standard real-world datasets demonstrate that for the datasets with fewer tags and according to the metrics sensitive only to highly ranked labels the seq2seq autoregressive models are consistently better than other models, even if we did not tune the label order for specific datasets.

In future work, we will provide more interpretation on the autoregressive and non-autoregressive architectures to unearth the reason of their strength and weaknesses and study more on the effect of the various label orders. We will also take advantage of the huge amount of partially labeled data to improve performance, and will investigate how well various methods cope with the missing labels.


\bibliographystyle{aaai21}
\bibliography{refs}


\end{document}